# An Independent Study of Reinforcement Learning and Autonomous Driving

Yang, Hanzhi
yanghz@umich.edu
University of Michigan, Mechanical Engineering

## Abstract

Reinforcement learning has become one of the most trending subjects in the recent decade. It has seen applications in various fields such as robot manipulations, autonomous driving, path planning, computer gaming, etc. We accomplished three tasks during the course of this project. Firstly, we studied the Q-learning algorithm [6] for tabular environments and applied it successfully to an OpenAi Gym environment [11], *Taxi* (see Appendix I). Secondly, we gained an understanding of and implemented the deep Q-network algorithm [6] for Cart-Pole environment (see Appendix II). Thirdly, we also studied the application of reinforcement learning in autonomous driving and its combination with safety check constraints (safety controllers). We trained a rough autonomous driving agent using *highway-gym* environment [9] and explored the effects of various environment configurations like reward functions on the agent training performance (see Appendix III).

# I. Reinforcement Learning and Q-learning

### i. Introduction

A reinforcement learning (RL) agent learns to make optimal decisions through a cost function by interacting with the environment directly. A RL problem is generally formulated as a Markov Decision Process (MDP) tuple $<S, A, T, R>$ in which there are four elements: state space $S \in \mathbb{R}^n$, action space $A$, state transformation function $T: S \times A \to S$, and reward function $R: S \times A \times S \to \mathbb{R}$. RL algorithms are mainly in two categories: value based, and policy based. In value-based methods, a value function is stored, and the policy is implicit and derived directly from the value function; on the contrary, in policy-based method, a representation of a policy, i.e., a mapping of state to action, is built and kept in memory during the learning process. In this report, we focused on value based RL algorithm, i.e., Q-learning methodology [6].

At each time step, the agent needs to select an action $a_t$ from its action space $A$ according to some policy $\pi$ based on the current state where $s_t$, $a_t = \pi(s_t)$. Applying this action leads the system state $s_t$ to the next state $s_{t+1}$, and the agent also receives a reward from the environment $r_t$. The agent will repeat such process until it has proceeded $N_s$ samples or it reaches the terminal state; thus ends a learning episode. After one episode, the agent restarts the whole learning process until $N_e$ episodes or policy convergence. In the Q-learning RL, all systems are considered to be an MDP [1].

The goal of Q-learning is to find a policy $\pi$ so that the total accumulated reward $R_t = \sum_{k=0}^{\infty} \gamma^k r_{t+k}$ can be maximized. In the total accumulated reward equation, the scalar constant term $\gamma \in (0,1]$ is the discount factor, meaning that the future reward received would decrease as the time step gets longer. The discount factor affects the agent's long-term decision making: for a discount factor closed to 1, the agent would tend to consider "further steps" as it may lead to more rewards, and for a small discount factor closed to 0, the agent would consider less steps as the rewards drop rapidly with time steps.

For discrete action space, the agent finds the optimal policy by solving a Q function, or action value function. The Q function satisfies the Bellman equation:

$$Q^*(s,a) = \mathbb{E}_{s'}[r + \gamma \cdot \max Q^*(s',a')|(s,a)] \quad \text{(Eq. 1.1)}$$

which leads to the Q value update function which eventually converges to the actual Q-table:

$$New\ Q(s,a) = Q(s,a) + \alpha[R(s,a) + \gamma \cdot \max Q'(s',a') - Q(s,a)] \quad \text{(Eq. 1.2)}$$

where the term $\alpha$ is learning rate and $\gamma$ is discount factor. After several iterations of learning, the agent would be able to make decisions by looking for any action $a$ that satisfies Eq.1.1 in state $s$ to be the optimal action in that state.

ii. *Experiment*

*Taxi-v3* is a toy text environment in the OpenAi Gym [11] package. In this environment, there are 4 locations labeled by different letters shown in figure 1.2.1 and the taxi agent needs to pick up a passenger at one location and drop the passenger off in another. The agent receives +20 points as rewards for a successful drop-off and loses 1 point for every timestep the whole process takes. Illegal pick-up and drop-off actions also lead to a -10 points penalty [2].

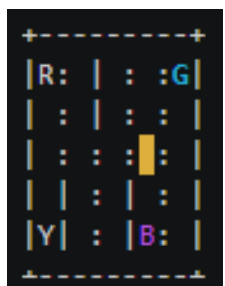

Figure 1.2.1. Taxi-v3 environment. Four locations are labels in letters R, Y, G, B. The yellow block indicates

the location of the taxi agent at the current step. The purple color letter represents the passenger location and the blue one, passenger's destination. The taxi can only pass through the dash lines while solid lines represent walls.

The state space of *Taxi-v3* includes all possible situations the taxi agent would inhabit. There exist 5 rows and 5 columns in the grid environment, and given 4 possible destinations and 5 passenger locations, the overall environment has $5 \cdot 5 \cdot 5 \cdot 4 = 500$ total possible states. The taxi agent encounters one of the 500 states and takes an action from the following six in the action space:

1. Going south
2. Going north
3. Going east
4. Going west
5. Pickup passenger
6. Dropoff passenger

In the learning process, the taxi agent iterates all the possible states and tries all actions for each of the states to upgrade its Q-table by the following steps:

1. Initialize the Q-table by all zeros
2. For each state, choose any one action from action space for the current state
3. Travel to the next state as a result of the action
4. For all possible actions from the state, select the one with the highest Q-value
5. Update Q-table values using equation 1.2
6. Set the next state as the current state
7. If the goal state is reached, end and repeat step 2~5

After several random explorations of actions, the Q-values would tend to converge serving the taxi agent as an action-value function which it can exploit to pick the most optimal action from a given state. To prevent the actions from always taking the same route and possibly overfitting, a parameter called tradeoff ($\epsilon$) is introduced to the algorithm, meaning that the agent does random exploration (exploration) occasionally with probability $\epsilon$ and takes the optimal action based on already learned Q-values (exploitation) most of the time with probability $1 - \epsilon$. In this experiment, we picked the following parameters:

Table 1.1. Taxi-v3 RL parameters

| Parameter | Value |
|---|---|
| Initial tradeoff | 1 |
| Final tradeoff | 0.1 |
| Tradeoff decay | 0.995 |
| Learning rate | 0.1 |
| Discount factor | 0.95 |

*iii.   Results and Discussion*

Using the parameters above and following the algorithm to update Q-table, the taxi agent was trained for 100,000 episodes. As shown in figure 1.3.1, the agent was able to reach the highest reward after around 600 episodes.

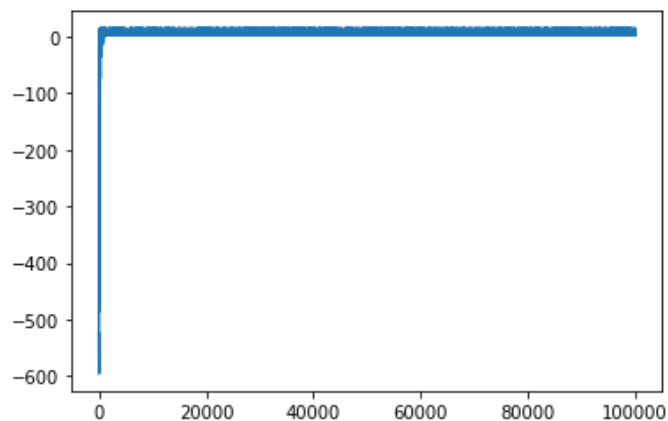

Figure 1.3.1. Taxi-v3 Q-learning reward curve

To test the performance of the agent, we randomly picked one state and print the Q-table at that state:

Table 1.2. Q-Table for state 328

|  | Action 1 | Action 2 | Action 3 | Action 4 | Action 5 | Action 6 |
|---|---|---|---|---|---|---|
| State 328 | -2.30 | -1.97 | -2.30 | -2.20 | -10.36 | -8.56 |

The maximum Q-value in for state 328 (see figure 1.3.2) is the second element -1.97, which means that the second action, going north, is the optimal action, which is logical because choosing this action the taxi is tending to approach the passenger at the northwest corner.

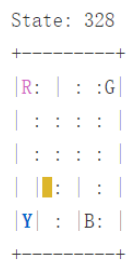

Figure 1.3.2. State 328. Passenger location is represented as purple R, and taxi as yellow block.

To further examine the efficiency of applying reinforcement learning in the taxi problem, we compared the Q-learning agent to a non-RL agent which randomly picks actions for each state until the environment target has been reached. We evaluate the two agents according to (1) average number of penalties per episode, (2) average number of timesteps per trip, and (3) average rewards per move. The results are shown in Table 1.3 below.

Table 1.3. RL agent and random agent comparisons over 100 episodes

|  | Q-learning agent | Random agent |
|---|---|---|
| average number of penalties per episode | 0 | 920.45 |
| average number of timesteps per trip | 12.38 | 2848.14 |
| average rewards per move | 0.69 | -3.91 |

As shown, the RL agent has all the three measures lower than those of the random agent. Hence, Q-learning is a simple and efficient way of reinforcement learning. However, a problem of Q-learning algorithm is that if the number of states in the environment is too high, the size of Q-table would be too large, and this makes the Q-table value iteration difficult and less efficient. To solve such problem, deep neural networks are introduced to reinforcement learning and upgrade Q-learning algorithm to Deep Q-learning Network (DQN) [6].

## II. Deep Reinforcement Learning and Deep Q-learning Network

### i. Introduction

*Curse of dimensionality* is a concept raised by Richard E. Bellman in 1957 [3]. This concept describes that when the dimensionality increases, the volume of the space increases so fast that the available data become sparse. This sparsity is problematic for any statistically significant methods. For example, 100 points can divide a unit domain by 0.01 sample-between distance, but if the dimensionality increases to 10, to achieve the 0.01 sample-between distance, $10^{20}$ points are needed, and thus the 10-dimension space is $10^{18}$ times larger than the unit domain. Such dimensional curse commonly exists in dynamic programming [3][4]. When there exist too many states, it is difficult to include all the state space in a table—even if all the states are included in a table, it is difficult to save and sweet the values. Therefore, the dimensionality of the state space needs to be compressed and the solution is *value function approximation*.

Instead of using a whole table of Q-values, value function approximation uses a function to represent $Q(s, a) = f(s, a)$, and $f$ can be any types of function, e.g., linear function $Q(s, a) = w_1 \cdot s + w_2 \cdot a + b$ in which $w1, w2, b$ are the linear coefficients. With this value function, no matter how large the dimensionality of state space $s$ is, the calculation of Q-value is through matrix calculations instead of one-by-one iterations. And since the distribution of Q-values is unknown, value function approximation uses the function $f$ to approximate the Q-values distribution, and adding $f$'s coefficients $w$ yields $Q(s, a) \approx f(s, a, w)$.

Deep learning (DL) is a machine learning algorithm which use multiple layers to progressively extract higher-level features from the raw input [5]. For instance, when dealing with an image with many pixel values, lower layers may identify edges and higher layers may identify the concepts relevant to an object target like a digit or a letter. Combining the concepts of deep learning and Q-learning algorithms gives the deep Q-learning network (DQN). However, there is a difference between deep learning and reinforcement learning that deep learning is a type of supervised learning algorithm which needs tagged samples to learn yet reinforcement learning is unsupervised and thus does not need any samples to learn. To fit reinforcement learning algorithms into a deep learning network, tagged samples need to be provided to the Q-learning network.

As shown in equation 1.1, the update of Q-values is based on reward and the target Q-value for next state. Hence, in DQN the target Q-values are used as its tagged samples. As a result, the DQN training loss function is

$$L(w) = \mathbb{E}[(r + \gamma \cdot \max Q(s', a', w) - Q(s, a, w))^2]  \quad \text{(Eq. 2.1)}$$

in which $s', a'$ are the next state and action and the highlight equation is the target Q-value. With the loss function, the DQN algorithm [6] can be solved as

**Algorithm 1** Deep Q-learning with Experience Replay
Initialize replay memory $\mathcal{D}$ to capacity $N$
Initialize action-value function $Q$ with random weights
**for** episode $= 1, M$ **do**
    Initialise sequence $s_1 = \{x_1\}$ and preprocessed sequenced $\phi_1 = \phi(s_1)$
    **for** $t = 1, T$ **do**
        With probability $\epsilon$ select a random action $a_t$
        otherwise select $a_t = \max_a Q^*(\phi(s_t), a; \theta)$
        Execute action $a_t$ in emulator and observe reward $r_t$ and image $x_{t+1}$
        Set $s_{t+1} = s_t, a_t, x_{t+1}$ and preprocess $\phi_{t+1} = \phi(s_{t+1})$
        Store transition $(\phi_t, a_t, r_t, \phi_{t+1})$ in $\mathcal{D}$
        Sample random minibatch of transitions $(\phi_j, a_j, r_j, \phi_{j+1})$ from $\mathcal{D}$
        Set $y_j = \begin{cases} r_j & \text{for terminal } \phi_{j+1} \\ r_j + \gamma \max_{a'} Q(\phi_{j+1}, a'; \theta) & \text{for non-terminal } \phi_{j+1} \end{cases}$
        Perform a gradient descent step on $(y_j - Q(\phi_j, a_j; \theta))^2$ according to equation 3
    **end for**
**end for**

As highlighted above, a major part of DQN algorithm is the experience replay. Assume the samples collected by the model is a continuous time sequence, then if the Q-values are updated every time a new sample is collected, the output would not be good due to the sample distribution. Therefore, the samples are firstly saved in a batch and then picked randomly; this is the process of experience replay, i.e., learning from the memories.

ii.   Experiment

*CartPole-v1* is a classic control environment in open-ai gym package. This environment shown in figure 2.2.1 simulates a commonly used state space feedback control problem, inverted pendulum (IP) [7]. In such problem, a pole is attached to a cart by an unactuated joint. The cart is able to move along a frictionless track of limited length. By controlling the movement of the cart, the system aims for balancing the pole upright and preventing it from falling over. In the environment, the cart agent receives +1 or -1 as its action to move left or right. It gets a +1 reward every timestep that the pole is upright. Each episode ends when the pole is more than 15 degrees from vertically upward, or the cart moves more than 2.4 units from the center of the track.

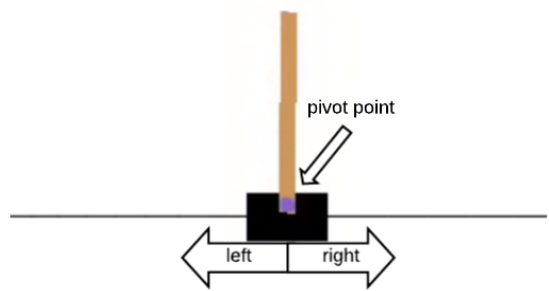

Figure 2.2.1. CartPole-v1 environment. Cart as the black block can move freely along the track, and the pole as the brown block can rotate accordingly around the purple pivot point. The goal is to balance the pole upright and prevent it from falling over.

Each of the states in the state space of the cart-pole environment includes the position and velocity of the pendulum and the cart, represented as a tuple with four elements: $< \theta, \dot{\theta}, x, \dot{x} >$. And every action in the action space includes the two possible movements of the cart: (1) moving to the left and (2) moving to the right. For the cart-pole environment, this is a high dimensional input (states of the IP system) and low dimensional output (discrete actions). Therefore, the value function approximation needs to compress the dimensionality of the state space $Q(s) = [Q(s, a_1), Q(s, a_2)] \approx f(s, w)$.

In the experiment, we used the following parameters to build our DQN:

Table 2.1. DQN parameters

| Parameter | Value |
| --- | --- |
| Discount Factor | 0.95 |
| Initial exploration rate | 1.0 |
| Final exploration rate | 0.001 |
| Decay | 0.995 |
| Batch size | 64 |

and we used 4 layers of neural works with sizes of 512, 256, 64, and 2. In the experience replay part, we used $Q(s, a; \theta)$ to represent the output of the current network, or Main Net (MN). The MN is used to evaluate the value function of the current state and action combination. After several iterations, the values in MN are copied to the target network (TN). TN enables the target Q-values to remain unchanged for a range of time, and thus lower the correlation of current Q-values and target Q-values. Hence, using such MN-TN method are we able to enhance the robustness and stability of the DQN algorithm.

### iii. Results and Discussion

We ran the DQN for 1000 episodes. As shown in figure 2.3.1, the IP agent is able to receive a high reward consistently after 300 episodes. Comparing to the traditional way of system control which requires a throughout system modeling and a state space feedback controller design using pole placement or LQR method, DQN algorithms makes it simpler to balance the

IP system without knowing any parameters of the system.

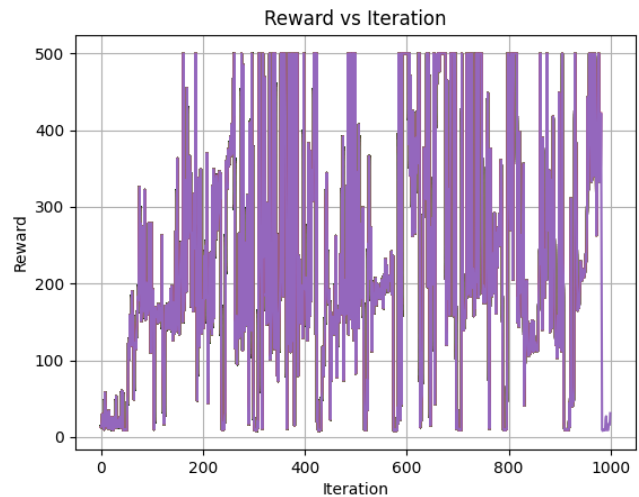

Figure 2.3.1. DQN reward curve.

The DQN algorithm on IP system, however, still has some limitations. Even though no IP system modeling is needed, the parameters of the DL neural network matter a lot for the agent training performance. Getting the right values of the hyperparameters needs a lot of experimentation. In addition, because of the need for the target Q-network to stabilize training and use of replay buffer to address catastrophic forgetting, the DQN agent is not trainable in an online manner. Another limitation is that the current DQN agent is very basic, so once the agent is applied to a different IP system, e.g., an IP with a different pole length or mass, the DQN agent would not be able to balance it. On the contrary, a state feedback controller would finish the task because the controller design allows the existence of such disturbance to the system.

The DRL method has been proven to be capable of solving complex tasks with huge dimensionality of state and action spaces. Certain methods have been developed to solve its limitations, such as adaptive reinforcement learning, multi-agent reinforcement learning, etc.

## III. Reinforcement Learning in Autonomous Driving

*i.    Introduction*

In 2020, there are over 2000 papers published in reinforcement learning and autonomous driving (AD) [8]. It appears that the application of RL in AD for continuous decision making has become one of the most trending AD areas. RL is commonly used in AD in the following aspects:

(1) Due to the complexity of motion planning, traditional control theory is not able to completely solve the problems, so reinforcement learning is widely used in this area. Yet, neural network is an end-to-end black box model, so most of the current studies tends to divide AD into several sub-problems like highway driving, crossing, parking, intersecting, etc., and solve each of them by different methods. For example, a rough AD procedure is: first, the road network data like user destination and online traffic information are used for route planning, and then with the route planning, the vehicle gets its waypoints, which are transferred to its behavioral layers controlling motions like car following, lane changing, and so on; with all the vehicle behavior strategy, the vehicle starts motion planning, estimating pose and collision free space, which leads to a trajectory, and this trajectory is passed to the vehicle's local feedback controller, which uses modern control theories to control vehicle states such as steering, throttle, and brake.

(2) All the RL theories are based on MDP, and AD is a kind of partial observable process (POMDP), i.e., the vehicle is able to observe the states of traffic and itself, but it lacks the capability of knowing future states of other road users like other vehicles or passengers.

(3) An ego vehicle is a car with environment sensors that can observe the environment states.

For a vehicle model, the precision of sensing and its own computing resources need to be balanced, using 2 dof, 3 dof, or 9 dof. For traffic environment, there already exist a lot of simulations, but since a simulation is not equal to the real world, the difference between it and real world would greatly affect the performance of RL in AD.

(4) To design a motion planning system based on RL, the first problem would be action space. There are some researchers separating decision making and control into two different problems, but some combine these two into a longitudinal and lateral behaviors control problem. Also, the reward function matters a lot in RL in AD. The rewards design greatly affects the resulting ego-vehicle behaviors like its target speed and collision avoidance. Next problem is the definition of state space. The observation of an ego-vehicle can be really large as the traffic model is very complicated, including the positions and speeds of other road users, traffic signal lights in front of the ego-vehicle, traffic signs on the side of the roads, etc. A common idea to solve the state space problem is to simplify the physical world by seeking the important information only. This can be divided into two perspectives: the first is computer vision, which uses CNN to find important information like car lanes, other vehicles, and so on, the other is grid matrix, which views the environment from a bird's eye view and represent every state of other road users around the ego-vehicle in a hyper grid matrix.

ii. *Experiment*

*Highway-env* is an autonomous driving simulation environment developed by Edouard Leurent in 2018 [9]. In this environment shown in figure 3.2.1, the ego-vehicle is driving on a multilane highway surrounding with other vehicles. The ego-vehicle agent's goal is to reach a high speed while avoiding collisions with neighboring vehicle and keeping driving on the rightmost lane of the road. By default, the agent receives a $R(s,a) = a \cdot \frac{v - v_{min}}{v_{max} - v_{min}} - b \cdot collision$ reward in which $v, v_{min}, v_{max}$ are the current, minimum, and maximum speed of the agent and $a, b$ are 2 user-defined coefficients.

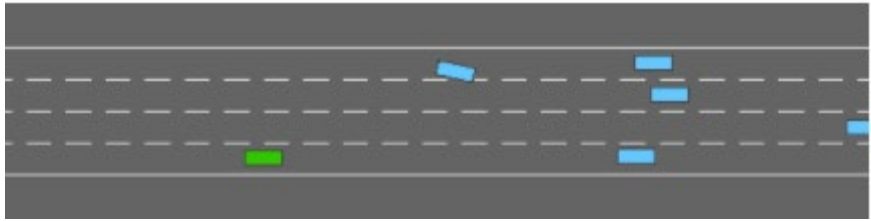

Figure 3.2.1. Highway environment. The ego-vehicle agent is represented as the green block and other vehicles on the road are represented as blue blocks. Blue vehicles are randomly generated in front of the agent and can change lanes randomly. The agent's objective is to avoid any kind of collisions with its neighboring vehicle while maintaining the maximum speed.

The state space of highway-env includes the longitudinal and lateral positions and velocities of every vehicle visible on the road. The action space includes 5 actions for the ego-vehicle including 1- moving to left lane, 2- idle, 3- moving to right lane, 4- moving faster, 5- moving slower. For the DQN agent training, we build the network using *Stable Baselines DQN* [10] and used the following parameters:

Table 3.1. DQN parameters

| Parameter | Value |
| --- | --- |
| Discount factor | 0.99 |
| Learning rate | 0.0005 |
| Buffer size | 50000 |
| Exploration fraction | 0.1 |
| Initial exploration rate | 1.0 |
| Final exploration rate | 0.02 |

| Batch size | 32 |

*iii.    Results and Discussion*

We firstly ran the DQN for 50,000 episodes, as shown in figure 3.3.1 as the orange baseline curve. We noticed the large time consumption of training a DQN model, so we altered the DQN into a double deep Q-learning network (DDQN) and trained this model for another 20000 episodes. Shown as the blue curve in the figure, DDQN takes less time to receive a higher reward than the baseline. So, we then tested this model in a 4-lane highway simulation. As tested in the environment, the agent seems to have the "awareness" to avoid the vehicle in front of it while speeding up. Yet, in some certain circumstances, the vehicle speeds up and avoids the first vehicle in front of it, and then since there is one vehicle in front and one on the left lane, the best decision a human driver would make would be taking the right lane, but the agent chooses the left lane and thus collides with the left lane vehicle. We think problems as such are caused by the lack of long-term reward received by the agent during the training.

Therefore, we further altered the reward function of the agent by increasing the penalty of collision from -1 to -5 and decreasing the reward of high speed from [20,30] to [15,20]. We trained the new model for 10000 episodes and found that, shown as the red curve in the figure, the agent receives high rewards comparing to the previous two models. Hence, we tested the effects of RL algorithms, DQN parameters, and reward functions on the ego-vehicle performance. However, we also found that although the third agent has the largest reward function, sometimes to avoid collision it keeps at a very low speed, and this is not what the agent is targeted to achieve. Therefore, further balancing between the weights of collusion avoidance and high-speed maintaining is required for agent's better performance.

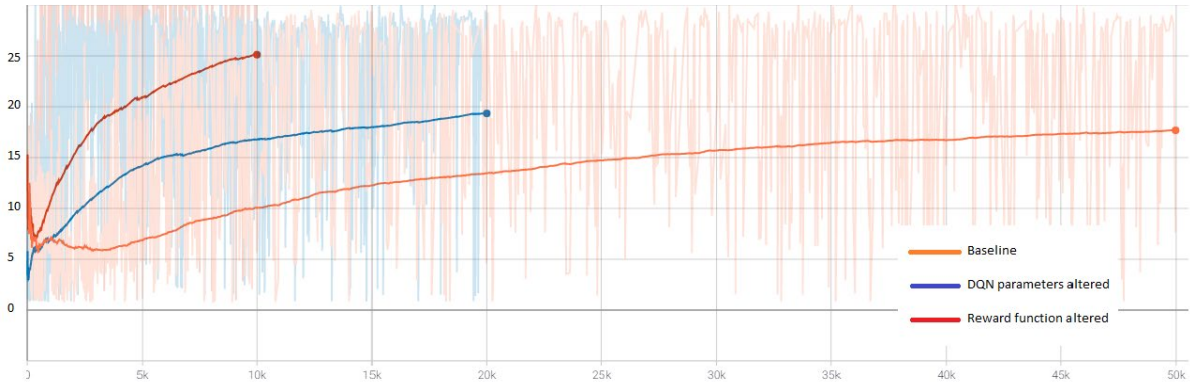

Figure 3.3.1. DDQN reward curves.

DRL has been proved to have the capability of solving many AD sub-problems, but it still has a lot of limitations. First, to actually apply the RL technology to real-world vehicles, technology standards are required in this area. Second, DRL need a lot of training and repeated model optimizations to solve complex problems. Third, to design a robust RL system, the difference gap between simulations and real world needs to be eliminated—such gap could hardly be solved by simply adding noises to the simulations. We believe that a combination of RL and supervised learning can be a key to solve more AD problems. For example, in the car following sub-problem, considering the comfortableness of passengers in such human-machine interaction environment, certain human driver decision making data can be collected so that the ego-vehicle can be trained in a model-based method. In some complex traffic environment, DRL can be trained in simulation using lots of real human driver dataset and then find even better solutions using RL.

## Conclusion

In this report, we have studied Q-learning algorithm in reinforcement learning and its combination with deep learning algorithm as deep Q-learning network. We conducted two experiments, taxi and cartpole, to apply these algorithms to solve real problems. Furthermore, we studied the applications of deep reinforcement learning in autonomous driving and conducted an experiment to examine the influence of Q-learning network parameters and agent reward functions on the reinforcement learning performance. The results show that DRL is a potentially powerful tool to solve complicated real-world issues but still has a lot of limitations. To solve these limitations, either advanced RL algorithms or a combination of model-based deep learning and reinforcement learning is needed.

# Appendix I: Taxi Agent Python Codes

'''
**Initialize environment.**
'''
import gym
env = gym.make("Taxi-v3").env
env.render()

'''
**test car position (4,2), passenger index 3, drop off location 2**
'''
state = env.encode(4, 2, 3, 2)
# using this as starting point
print("State: ", state)
env.s = state
env.render()
# as shown, the current position is defined as state 454
env.P[env.s]

'''
**use Q-learning**
'''
import random
import numpy as np

# initialize Q table
q_table = np.zeros((env.observation_space.n, env.action_space.n))
q_table_old = np.ones((env.observation_space.n, env.action_space.n)) * -np.Inf
q_table_new = np.ones((env.observation_space.n, env.action_space.n)) * np.Inf

# setting parameters
learning_rate = 0.1
discount_factor = 0.95
trade_off = 1
decay = 0.995
# i = 0
diff = 1

for i in range(1000):
# while True:

```python
        state = env.reset()
        # state = env.encode(4, 2, 3, 2)
        # env.s = state
        # env.render()
        # print(i)

        done = False
        while not done:
            if np.random.rand() <= trade_off:
                action = env.action_space.sample() # explore
            else:
                action = np.argmax(q_table[state]) # exploit

            next_state, reward, done, info = env.step(action)

            q_old = q_table[state, action]
            next_max = np.max(q_table[next_state])
            q_new = q_old + learning_rate * (reward + discount_factor * next_max - q_old)
            q_table[state, action] = q_new

            state = next_state

        if i > 200:
            trade_off *= decay
        if i > 5:
            q_table_old = q_table_new
            q_table_new = q_table
            diff = np.max(np.abs(q_table_new-q_table_old))

        # i += 1

        if diff < 0.000000000000001:
            # break
            pass

print(q_table)
print(diff)

'''
```
**Test agent**
```
'''
ACTION = ["DOWN", "UP", "RIGHT", "LEFT", "PICK", "DROP"]
```

```python
env.s = 454
print(np.argmax(q_table[env.s]), ":", ACTION[np.argmax(q_table[env.s])])
print(env.s)
env.render()

env.s = 354
print(np.argmax(q_table[env.s]), ":", ACTION[np.argmax(q_table[env.s])])
print(env.s)
env.render()

env.s = 254
print(np.argmax(q_table[env.s]), ":", ACTION[np.argmax(q_table[env.s])])
print(env.s)
env.render()

env.s = 274
print(np.argmax(q_table[env.s]), ":", ACTION[np.argmax(q_table[env.s])])
print(env.s)
env.render()

env.s = 374
print(np.argmax(q_table[env.s]), ":", ACTION[np.argmax(q_table[env.s])])
print(env.s)
env.render()

env.s = 474
print(np.argmax(q_table[env.s]), ":", ACTION[np.argmax(q_table[env.s])])
print(env.s)
env.render()
```

# Appendix II: Cart-Pole Agent Python Codes

```python
import numpy as np
import matplotlib.pyplot as plt
import gym
import random

from keras.models import Sequential, load_model
from keras.layers import Dense
from keras.optimizers import Adam
from collections import deque

"""
Constants
"""

ENV_NAME = "CartPole-v1"
ENV = gym.make(ENV_NAME)
EPISODES = 1000
DISCOUNT_RATE = 0.95
EXPLORATION_RATE = 1.0
MIN_EXPLORATION_RATE = 0.001
DECAY = 0.995
BATCH_SIZE = 64

"""
Model Initialization
"""

def Build_Model(shape, action_space):
    model = Sequential()

    model.add(Dense(512, input_shape = shape, activation = "relu"))
    model.add(Dense(256, activation = "relu"))
    model.add(Dense(64, activation = "relu"))
    model.add(Dense(action_space, activation = "linear"))

    model.compile(loss = "mse", optimizer = Adam(lr = 0.001))

    return model
```

"""
Deep Q-Learning Network
"""

```python
class DQN:

    def __init__(self):
        self.env = ENV
        self.state_num = self.env.observation_space.shape[0]
        self.action_num = self.env.action_space.n

        self.episodes = EPISODES
        self.memory = deque(maxlen = 2000)

        self.gamma = DISCOUNT_RATE
        self.epsilon = EXPLORATION_RATE
        self.min_epsilon = MIN_EXPLORATION_RATE
        self.decay = DECAY
        self.batch_size = BATCH_SIZE

        self.model = Build_Model((self.state_num,),self.action_num)

    def remember(self, state, action, reward, next_state, done):
        self.memory.append((state, action, reward, next_state, done))

    def act(self, state):
        if np.random.rand() <= self.epsilon:
            # explore
            action = random.randrange(self.action_num)
        else:
            # exploit
            q_values = self.model.predict(state)
            action = np.argmax(q_values)
        return action

    def replay(self):
        if len(self.memory) < 1000:
            return
        batch = random.sample(self.memory, self.batch_size)
```

```python
        # batch[i] = [state, action, reward, next_state, done]
        state = np.zeros((self.batch_size, self.state_num))
        next_state = np.zeros((self.batch_size, self.state_num))
        action, reward, done = [], [], []

        for i in range(self.batch_size):
            state[i] = batch[i][0]
            action.append(batch[i][1])
            reward.append(batch[i][2])
            next_state[i] = batch[i][3]
            done.append(batch[i][4])

        q = self.model.predict(state)
        q_new = self.model.predict(next_state)

        for i in range(self.batch_size):
            if done[i]:
                q[i][action[i]] = reward[i]
            else:
                q[i][action[i]] = reward[i] + self.gamma * (np.amax(q_new[i]))

        self.model.fit(state, q, verbose = 0, batch_size = self.batch_size)

        self.epsilon *= self.decay
        self.epsilon = max(self.epsilon,self.min_epsilon)

    def train(self):
        t = []
        r = []

        def plot(x,y,n):
            plt.plot(x,y)
            plt.xlabel("Iteration")
            plt.ylabel("Reward")
            plt.grid(True)
            plt.title("Reward vs Iteration")
            name = "Figure-" + str(n) + ".png"
            plt.savefig(name)

        for ite in range(self.episodes):
            print(ite)

            state = self.env.reset()
```

```python
            state = state.reshape((1,self.state_num))

            done = False
            step = 0

            while not done:
                self.env.render()

                action = self.act(state)
                next_state, reward, done, _ = self.env.step(action)
                next_state = next_state.reshape((1,self.state_num))

                if not done or step == self.env._max_episode_steps - 1:
                    reward = reward
                else:
                    reward = -100

                self.remember(state, action, reward, next_state, done)
                state = next_state
                step += 1

                if done:
                    print("iteration: ", ite)
                    print("reward: ", step)
                    t.append(ite)
                    r.append(step)
                    if step >= 500:
                        print("Saving trained model as cartpole-dqn.h5")
                        self.model.save("cartpole-dqn.h5")
                        # return
                    break

                self.replay()

            if ite % 200 == 0 and ite >= 200:
                plot(np.array(t),np.array(r),ite/200)

t = np.array(t)
r = np.array(r)
plot(t,r,1)
print("Saving trained model as cartpole-dqn.h5")
# self.model.save("cartpole-dqn.h5")
```

```python
def cont_train(self):
    self.model = load_model("cartpole-dqn.h5")
    t = []
    r = []

    def plot(x,y,n):
        plt.plot(x,y)
        plt.xlabel("Iteration")
        plt.ylabel("Reward")
        plt.grid(True)
        plt.title("Reward vs Iteration")
        name = "Figure-" + str(n) + ".png"
        plt.savefig(name)

    for ite in range(100):
        print(ite)

        state = self.env.reset()
        state = state.reshape((1,self.state_num))

        done = False
        step = 0

        while not done:
            self.env.render()

            action = self.act(state)
            next_state, reward, done, _ = self.env.step(action)
            next_state = next_state.reshape((1,self.state_num))

            if not done or step == self.env._max_episode_steps - 1:
                reward = reward
            else:
                reward = -100

            self.remember(state, action, reward, next_state, done)
            state = next_state
            step += 1

            if done:
                print("iteration: ", ite)
                print("reward: ", step)
                t.append(ite)
                r.append(step)
```

```python
                    if step >= 500 and ite >= 50:
                        print("Saving trained model as cartpole-dqn.h5")
                        self.model.save("cartpole-dqn.h5")
                        t = np.array(t)
                        r = np.array(r)
                        plot(t,r,2)
                        return
                    break

            self.replay()

        t = np.array(t)
        r = np.array(r)
        plot(t,r,2)
        # print("Saving trained model as cartpole-dqn.h5")

    def test(self):
        self.model = load_model("cartpole-dqn.h5")
        for e in range(10):
            state = self.env.reset()
            state = state.reshape((1, self.state_num))
            done = False
            i = 0
            while not done:
                self.env.render()
                action = np.argmax(self.model.predict(state))
                next_state, _, done, _ = self.env.step(action)
                state = next_state.reshape((1, self.state_num))
                i += 1
                if done:
                    print("episode: {}, score: {}".format(e, i))
                    break

"""
Main
"""
if __name__ == "__main__":
    Inv_Pend = DQN()

    # Inv_Pend.train()
```

```
# Inv_Pend.cont_train()

Inv_Pend.test()
```

# Appendix III: Highway-env Python Codes

## 1. Agent training

```
import gym
import highway_env
import numpy as np

from stable_baselines import HER, SAC, DDPG, TD3
from stable_baselines.ddpg import NormalActionNoise

from stable_baselines.common.vec_env import DummyVecEnv
from stable_baselines.deepq.policies import MlpPolicy, CnnPolicy
from stable_baselines import DQN

env = gym.make("highway-v0")
ENV_NAME = 'dqn_highway_4-1(2)'
ENV_NAME_SAVE = 'dqn_highway_4-1(3)'

# Environment Configuration
env.config["lanes_count"] = 4
env.config['collision_reward'] = -5

env.reset()
model = DQN.load(ENV_NAME, env=env, verbose =1,
      tensorboard_log="./dqn_highway_tensorboard/", learning_starts=100)

print("Training start")
model.learn(total_timesteps=10000, tb_log_name="4-1")
model.save(ENV_NAME_SAVE)
print("Model saved")

del model

# Test agent model
model = DQN.load(ENV_NAME_SAVE, env=env)
obs = env.reset()

# Evaluate the agent
episode_reward = 0
for _ in range(500):
    action, _ = model.predict(obs)
    obs, reward, done, info = env.step(action)
```

```
        env.render()
        episode_reward += reward
        if done or info.get('is_success', False):
            print("Reward:", episode_reward, "Success?", info.get('is_success', False))
            episode_reward = 0.0
            obs = env.reset()
```

## 2. Agent testing for different environment configurations

```
import gym
import highway_env
import numpy as np

from stable_baselines.common.vec_env import DummyVecEnv
from stable_baselines.deepq.policies import MlpPolicy, CnnPolicy
from stable_baselines import DQN

from stable_baselines.common.vec_env import VecVideoRecorder

ENV_NAME = 'dqn_highway_4-1(2)'

env = gym.make("highway-v0")
env.config['screen_height'] = 300
env.config["lanes_count"] = 3
env.config["duration"] = 1000
env.config["show_trajectories"] = False
env.config["vehicles_density"] = 1
env.config["vehicles_count"] = 50

model = DQN.load(ENV_NAME, env=env)
obs = env.reset()

# Evaluate the agent
episode_reward = 0
for _ in range(1000):
    action, _ = model.predict(obs)
    obs, reward, done, info = env.step(action)
    env.render()
    episode_reward += reward
    if done or info.get('is_success', False):
        print("Reward:", episode_reward, "Success?", info.get('is_success', False))
        episode_reward = 0.0
        obs = env.reset()
```